\documentclass[10pt,twocolumn,letterpaper]{article}

\usepackage{cvpr}              
\usepackage{cvpr}              
\usepackage{graphicx}
\usepackage{amsmath}
\usepackage{amssymb}
\usepackage{booktabs}
\usepackage{appendix}
\usepackage{multicol}
\usepackage{amsfonts}
\usepackage{bibentry}
\usepackage{multirow}
\usepackage{amssymb}
\usepackage{pifont}
\newcommand{\cmark}{\ding{51}}%
\newcommand{\xmark}{\ding{55}}%
\usepackage{bbding}
\usepackage[accsupp]{axessibility}

\input{def.set}

\usepackage[pagebackref,breaklinks,colorlinks]{hyperref}

\newcommand{\tablestyle}[2]{\setlength{\tabcolsep}{#1}\renewcommand{\arraystretch}{#2}\centering\footnotesize}

\usepackage[capitalize]{cleveref}
\crefname{section}{Sec.}{Secs.}
\Crefname{section}{Section}{Sections}
\Crefname{table}{Table}{Tables}
\crefname{table}{Tab.}{Tabs.}

\def\cvprPaperID{3738} 
\def\confName{CVPR}
\def\confYear{2022}
\usepackage[english]{babel}

\begin{document}

\title{RU-Net: Regularized Unrolling Network for Scene Graph Generation}

\author{Xin Lin$^1$\thanks{Work done during first author’s internship at JD Explore Academy.} \quad Changxing Ding$^{1,2}$\thanks{Corresponding author.}  \quad Jing Zhang$^3$ \quad Yibing Zhan$^{4}$ \quad Dacheng Tao$^{4,3}$\\
$^1$ South China University of Technology \quad
$^2$ Pazhou Lab, Guangzhou \quad
$^3$ The University of Sydney \quad \\
$^4$ JD Explore Academy\\
{{\tt\small eelinxin@mail.scut.edu.cn,\quad chxding@scut.edu.cn,\quad
jing.zhang1@sydney.edu.au,\quad}} \\
{{\tt\small zhanyibing@jd.com,\quad
dacheng.tao@gmail.com}}}
\maketitle

\begin{abstract}

Scene graph generation (SGG) aims to detect objects and predict the relationships between each pair of objects. Existing SGG methods usually suffer from several issues, including 1) ambiguous object representations, as graph neural network-based message passing (GMP) modules are typically sensitive to spurious inter-node correlations, and 2) low diversity in relationship predictions due to severe class imbalance and a large number of missing annotations. To address both problems, in this paper, we propose a regularized unrolling network (RU-Net). We first study the relation between GMP and graph Laplacian denoising (GLD) from the perspective of the unrolling technique, determining that GMP can be formulated as a solver for GLD. Based on this observation, we propose an unrolled message passing module and introduce an $\ell_p$-based graph regularization to suppress spurious connections between nodes. Second, we propose a group diversity enhancement module that promotes the prediction diversity of relationships via rank maximization.  Systematic experiments demonstrate that RU-Net is effective under a variety of settings and metrics. Furthermore, RU-Net achieves new state-of-the-arts on three popular databases: VG, VRD, and OI. Code is available at \href{https://github.com/siml3/RU-Net}{
https://github.com/siml3/RU-Net}.

\end{abstract}

\section{Introduction}
Scene Graph Generation (SGG) aims to provide a graphical representation of objects and their relationships in an image. Recently, SGG has emerged as a promising approach that bridges the gap between vision and natural language domains. It has been found to be useful for many vision tasks, including 3D scene understanding \cite{armeni20193d,wald2020learning}, visual question answering \cite{damodaran2021understanding,shi2019explainable}, and image captioning \cite{gu2019unpaired,zhang2020empowering}.

A scene graph comprises a collection of triplets in the form \textit{subject-relationship-object}. The objects and their pairwise relationships are denoted as nodes and edges, respectively.
Existing SGG models \cite{lin2020gps,yang2021probabilistic,li2021bipartite,zhong2021learning,tang2019learning,chen2019knowledge,yang2018graph} typically utilize context modeling strategies to learn discriminative representations for node and edge prediction; specifically, most of them adopt graph neural network-based message passing (GMP) mechanisms. In GMP, node representations are iteratively updated through the aggregation of neighboring information according to learnable attention weights, which are typically supervised by node labels.

\begin{figure}[t]
\begin{center}
    \vspace{-3.5mm}
    \centering
    \includegraphics[scale=0.53]{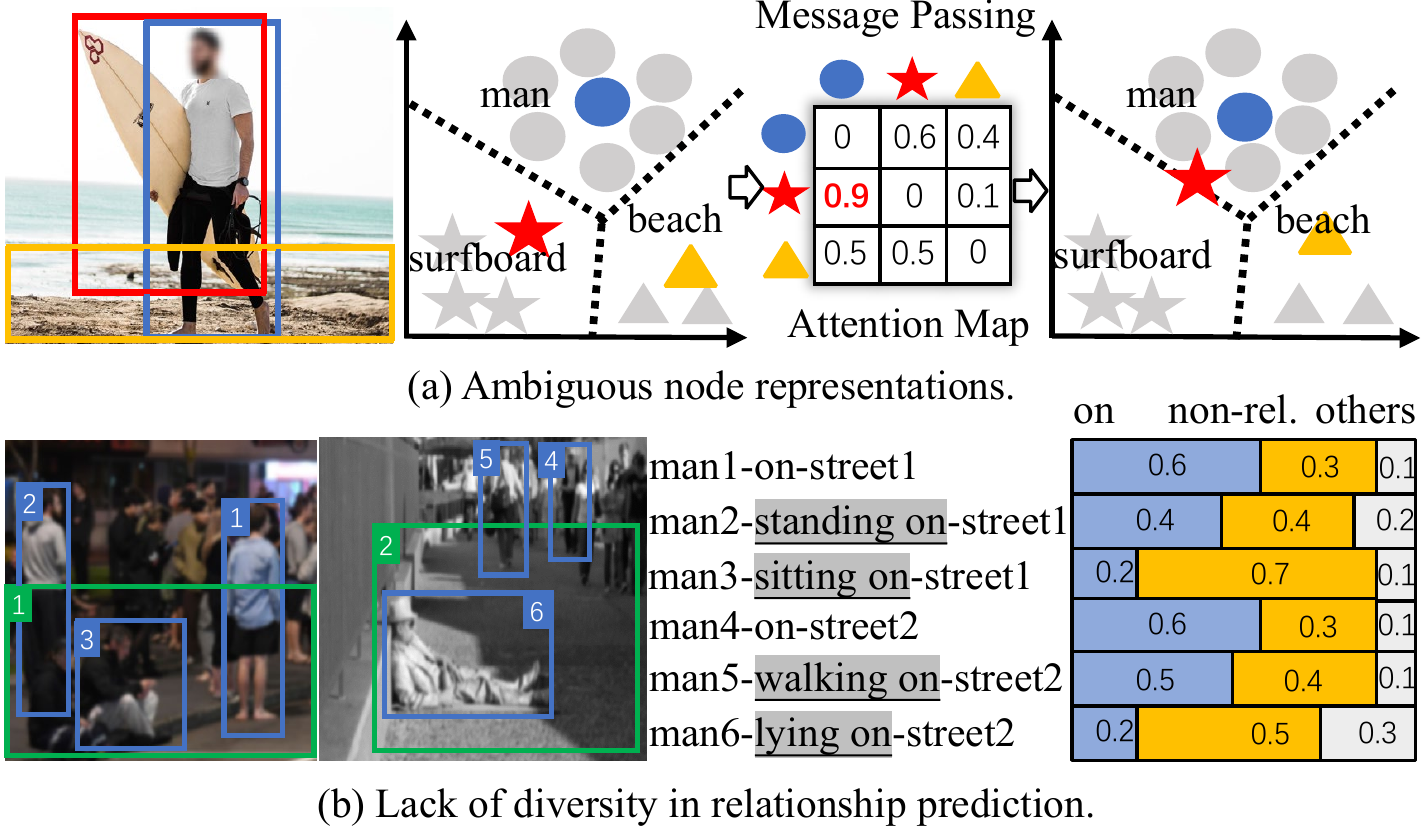}
    \vspace{-3mm}
    \captionof{figure}{(a) Spurious correlation between nodes causes ambiguous representations through graph neural network-based message passing. (b) Relationship prediction for the same category of node pairs lacks diversity. Missing relationship annotations are underlined and highlighted in gray. Best viewed in color.}\label{intro1}
     \vspace{-7mm}
\end{center}
\end{figure}

However, current GMPs are negatively impacted by spurious correlations between nodes. Here, a spurious correlation refers to a relatively large attention weight between a pair of semantically disparate nodes. These spurious correlations frequently occur, as attention weights between spatially proximate nodes tend to be large regardless of whether their object categories are related. In Figure~\ref{intro1}(a), it is evident that the attention weights for the {\textit{surfboard}} are dominated by those for the {\textit{man}} (\ie, equal to 0.9). As a result, the quality of representations for some nodes may degrade after erroneous message passing. Moreover, relationship prediction diversity among existing SGG models tends to be low. This is mainly due to the long-tailed distribution of relationships and a large number of missing relationship annotations. As shown in Figure~\ref{intro1}(b), the two images contain six triplets related to the {\textit{man-street}} pair; however, only two of them are annotated, and the relationship categories are both {\textit{on}}. The trained SGG models, therefore, tend to make biased predictions for the majority classes and the {\textit{non-relationship}} category.

To address the above issues, we propose a regularized unrolling network (RU-Net) for SGG. First, we study the relation between GMP and graph Laplacian denoising (GLD) \cite{shuman2013emerging} from the perspective of the unrolling technique \cite{monga2021algorithm}. We show that 1) GMP can be formulated as the solver for GLD, and 2) the quadratic penalty widely adopted in the formulation of GLD is sensitive to outliers (\eg, spurious correlations between nodes). As an alternative, we propose an unrolled message passing (U-MP) module and employ an $\ell_p$-based graph regularization term to suppress these spurious connections between nodes, thereby effectively reducing the ambiguity in node representations. Moreover, we determine that the optimization of the $\ell_p$-based graph regularization can be efficiently achieved in an end-to-end manner by integrating a reweighting matrix into U-MP, which accounts for the semantic dissimilarity between nodes.

Second, we introduce a group diversity enhancement (GDE) module to promote the diversity of relationship predictions for both labeled and unlabeled samples. More specifically, since score vectors for relationships tend to be linearly independent when predicted as different categories, we formulate the optimization of relationship prediction diversity as a rank maximization problem. Because rank maximization is NP-hard \cite{sprechmann2015learning}, we use the $\ell_{2,1}$-norm to approximate the matrix rank. We also divide the large matrix into several smaller ones, each of which contains relationship predictions for node pairs of the same object categories. By enlarging the $\ell_{2,1}$-norm of the smaller matrices, the relationship prediction diversity is more effectively optimized, as demonstrated in Section~\ref{ablation_all}.


In summary, the contributions of this study are three fold: (1) a novel unrolling framework that interprets GMP as a solver for the GLD problem; (2) the U-MP module for spuriousness-robust message passing via an $\ell_p$-based graph regularization, which enhances GMP's robustness against spurious connections between nodes; and (3) the GDE module, which improves the diversity of relationship prediction via the group-wise $\ell_{2,1}$-based regularization term. The efficacy of the proposed RU-Net is systematically evaluated on three popular SGG databases: Visual Genome (VG) \cite{krishna2017visual}, OpenImages (OI) \cite{kuznetsova2020open}, and Visual Relationship Detection (VRD) \cite{lu2016visual}. Experimental results show that our RU-Net consistently outperforms state-of-the-art methods.

\begin{figure*}[t]
\begin{center}
\includegraphics[scale=0.5]{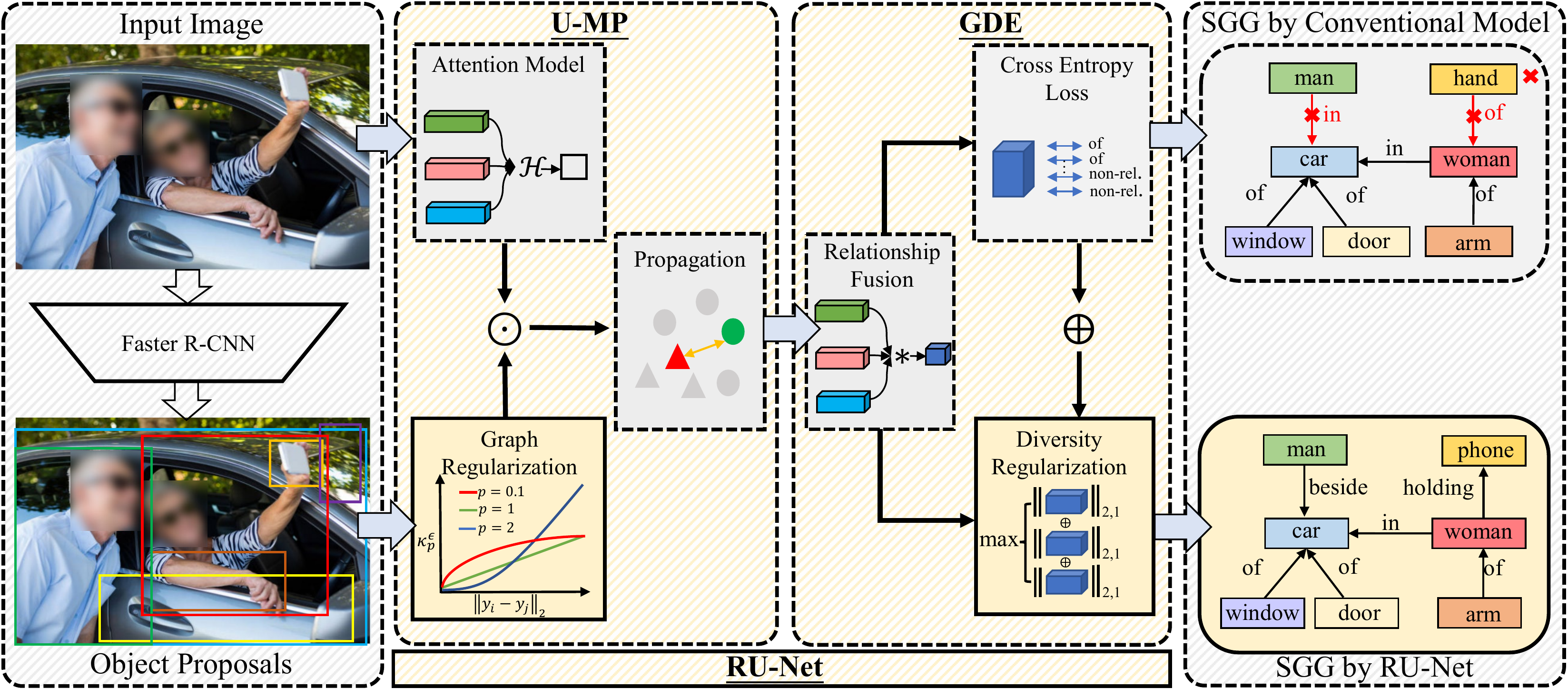}
\caption{The framework of RU-Net. RU-Net adopts Faster R-CNN \cite{ren2015faster} to obtain object proposals. Compared with conventional SGG models (highlighted in gray), our RU-Net promotes SGG model optimization with two regularization terms (highlighted in yellow). More specifically, the graph regularization acts as a reweighting matrix to refine the attention maps and reduce ambiguity in the node representations. The diversity regularization is incorporated with the cross-entropy loss and prompts the relationship prediction diversity via rank maximization. $\oplus$ and $\odot$ represent addition and the Hadamard product, respectively. The functions $\calH$ and $*$ are defined in Section~\ref{calH} and Section~\ref{GDE}, respectively. Best viewed in color.}
\label{frame}
\end{center}
\end{figure*}

\section{Related Work}
\textbf{Scene Graph Generation.} Existing works in SGG \cite{xu2017scene,yang2018graph,herzig2018mapping,chen2019knowledge,zhang2019graphical,chiou2021recovering,zhong2021learning} generally focus on context modeling or tackling the class imbalance problem (\ie, the long-tailed distribution). Several context modeling strategies have been proposed to learn discriminative object representation by exploring various message passing mechanisms. Zeller \etal \cite{zellers2018neural} represented the global context via a recurrent sequential architecture (\ie, bidirectional long short-term memory (Bi-LSTM) model). Tang \etal \cite{tang2019learning} utilized dynamic tree structures to realize node-specific message passing. Lin \etal \cite{lin2020gps} proposed a direction-aware message passing module that encodes the edge direction information into context modeling. Li \etal \cite{li2021bipartite} adopted a relationship prediction confidence-based adaptive message passing strategy to reduce noise in context modeling. Lu \etal \cite{lu2021context} utilized the transformer encoder to acquire contextual information pertaining to both objects and context. To handle the class imbalance issue, Tang \etal \cite{tang2020unbiased} proposed an unbiased model that removes the vision-agnostic bias with counterfactual causality, while \cite{chen2019soft,chiou2021recovering} addressed this problem using positive-unlabeled learning. Some works have additionally explored class imbalance learning strategies \cite{li2021bipartite,yan2020pcpl}, re-sampling and cost-sensitive learning, to relieve the long-tailed distribution problem. Our work considers both issues discussed above in a unified framework.

\textbf{Deep Algorithm Unrolling.}
In deep algorithm unrolling (DAU), the structure of the model-based iterative optimization algorithms is unrolled into a neural network \cite{mardani2018neural,monga2021algorithm,gregor2010learning}. Specifically, each iteration of the algorithm is represented as one layer of a network. Stacking these layers forms a deep neural network with an architectural structure that depends on the optimization method employed. The forward propagation of the network is equivalent to executing the iterative algorithm several times. Compared with fully parameterized neural networks, DAU is advantageous in terms of its interpretability and model complexity \cite{chen2021graph, li2020efficient,markowitz2021multimodal}; hence, DAU-based networks can be effectively optimized with less training data. For example, Yang \etal \cite{yang2018admm} proposed an unrolled version of the Alternating Direction Method of Multipliers \cite{yang2016deep} for magnetic resonance imaging. Zhang \etal \cite{zhang2018ista} integrated convolutional networks with the iterative shrinkage-thresholding algorithm \cite{beck2009fast} for compressed image sensing. Moreover, the half-quadratic splitting algorithm \cite{afonso2010fast} has been used in \cite{dong2018denoising, zhang2020deep} to unfold the minimization problems for image denoising and super-resolution. Inspired by these works, we introduce DAU to the SGG and unify existing GMP modules to solve GLD.

\section{Regularized Unrolling Network}
This section presents the details of the proposed regularized unrolling network. More specifically, we first introduce the preliminaries, then explain the network details and the training losses. As Figure~\ref{frame} illustrates, RU-Net comprises a U-MP module and a GDE module. From the perspective of DAU, the U-MP module utilizes $\ell_p$-based graph regularization to improve the robustness of existing GMP modules against spurious connections between nodes. For its part, the GDE module improves relationship prediction diversity via a group-wise $\ell_{2,1}$-based regularization term. In the below, we will describe these two components sequentially.

\subsection{Preliminaries}
\noindent \textbf{Notations}. To obtain the appearance feature for each proposal, we adopt the same approach used in \cite{zellers2018neural}. There are $O$ object categories (including background) and $R$ relationship categories (including non-relationship). The representation for the $i$-th node is denoted as ${\emph {\textbf{x}}_i}\in {\mathbb R}^{d}$. Specifically, ${\emph {\textbf{x}}_i}$ is obtained via linear projection from the concatenation of the appearance feature, object classification probabilities, and the spatial feature. For an image that includes $n$ nodes, we can obtain a node representation matrix $\bX\in {\mathbb R}^{n \times d}$, where $d$ is the feature dimension. In addition, we extract features from the union box of one pair of nodes $i$ and $j$, denoted as $\bu_{ij} \in {\mathbb R}^{d}$. $|\cdot|$, $\|\cdot\|_{2}$, and $\|\cdot\|_{F}$ denote the absolute value of a number, the $\ell_{2}$-norm of a vector, and the Frobenius norm of a matrix, respectively. $[;]$ represents the concatenation operation. $\odot$ is the Hadamard product. For a matrix $\bS \in \mathbb{R}^{m \times n}$, $[\bS]_{i j}$ and $\bs_{i}$ represent the $ij$-th entry and the $i$-th row of $\bS$, respectively.

\noindent \textbf{Smoothed $\ell_p$-norm Distance Metric}. To improve the robustness against spurious correlations between nodes, we utilize a smoothed $\ell_{p}$-norm distance metric \cite{song2015sparse} as follows:
\begin{equation}\label{smooth_p_ori}
\kappa_{p}^{\epsilon}(x) \triangleq \begin{cases} \epsilon^{p-2}|x|^{2}, & |x| \leq \epsilon \\ \frac{2}{p}|x|^{p}-\frac{2-p}{p} \epsilon^{p}, & |x|>\epsilon \end{cases} ,
\end{equation}
where $\epsilon>0$ and $0< p \leq 2$. As depicted in the coordinate plane of Figure~\ref{frame}, with a smaller value of $p$ (\eg, $p=0.1$), Eq.~\eqref{smooth_p_ori} places far less emphasis on large $|x|$ and is more robust against outliers than the $\ell_2$-based distance function. More details regarding the properties of Eq.~\eqref{smooth_p_ori} can be found in Appendix A.

\subsection{Unrolled Message Passing}
Existing SGG methods \cite{yang2018graph,liu2019knowledge,li2021bipartite,lin2020gps} typically utilize a sequence of GMP layers to iteratively refine node representations with contextual information. However, these GMP modules may be sensitive to spurious correlations between nodes, which may lead to more ambiguous node representations. To clarify and address this issue, we will discuss two key aspects in what follows: the relation between GMP and GLD \cite{shuman2013emerging} and spuriousness-robust graph regularization.

\subsubsection{The Relation between GMP and GLD}\label{calH}
In each GMP layer, a function is utilized to compute the attention weight for each node pair. The node representation is then updated by aggregating neighboring information according to the learnable attention weights. The output of the $k+1$-th GMP layer can be represented as follows:
\begin{equation}\label{mp-ori}
\left\{\begin{array}{l}
\bA^{(k+1)} = {\rm{Normalize}}(\calH(\bY^{(k)}))\\[5pt]
\bY^{(k+1)} ={\rm{ReLU}}(\bY^{(k)}+\bA^{(k+1)} \bY^{(k)})
\end{array}\right. ,
\end{equation}
where $\bY \in \mathbb{R}^{n\times d}$ represents the node representations refined by GMP. $\bA \in \mathbb{R}^{n \times n}$ stands for the learned attention matrix. $\calH(\bY)$ is a trainable attention function with $\bY$ as input. ``Normalize" denotes the row-wise normalization via softmax function.

Next, we will prove that the GMP module defined in Eq.~\eqref{mp-ori} essentially solves the GLD \cite{shuman2013emerging} problem in the SGG context. Specifically, the GLD problem can be defined as:
\begin{equation} \label{basic_F}
\calL_{_{\text{GLD}}}(\bY,\bL) \triangleq  \left\|\bY - \bX  \right\|_{F}^2 + \calG_{_{\text{GLR}}}(\bY,\bL) \ ,
\end{equation}
where
\begin{equation}
\label{eq:glr}
\calG_{_{\text{GLR}}}(\bY,\bL)=\left\| {\bL^{\frac{1}{2}}\bY} \right\|^2_{F} = \sum\nolimits_{(i,j)\in \calE} [\bA] _{ij}\left\| \by_{i} - \by_{j} \right\|_2^2.
\end{equation}Here, Eq.~\eqref{eq:glr} is well known as the Graph Laplacian Regularization (GLR) \cite{ortega2018graph}. $\calE$ denotes the entire set of node pairs in the scene graph. Unlike the standard GLD problem \cite{shuman2013emerging} where the Laplacian matrix $\bL$ is already known, this matrix needs to be learnt in SGG. Specifically, the Laplacian matrix is defined as follows: $\bL = \bD\!-\!\bA$, where $[\bD]_{ii} = \sum_j [\bA]_{ij}$. 

Motivated by the algorithm unrolling strategy \cite{monga2021algorithm}, we can unfold a sequence of gradient steps to form an unrolled message passing (U-MP) module and optimize Eq.~\eqref{basic_F}. Specifically, given $\bL$, we have
\begin{equation} 
\frac{\partial \calL_{_{\text{GLD}}}(\bY) }{\partial \bY} = 2\bL \bY + 2\bY - 2 \bY^{(0)},
\end{equation}
where $\bY^{(0)}=\bX$. Therefore, the $k+1$-th step in the gradient descent can be written as follows:
\begin{equation} \label{eq:basic_grad_step}
\bY^{(k+1)} = \bY^{(k)} - 2{\alpha}[ (\bL  + \bI) \bY^{(k)} - \bY^{(0)} ],
\end{equation}
where ${\alpha}$ is the step size and $\bI$ denotes an identity matrix. If we replace $\bL$ with the random-walk normalized Laplacian \cite{kipf2017semi} version $\bL=\bI-{\bD}^{-1}\bA$ and set $\alpha$ as 1/6, we have:
\begin{equation} \label{eq:basic_grad_step_sim}
\bY^{(k+1)} = \tfrac{1}{3}(\bD^{-1}\bA \bY^{(k)} + \bY^{(k)} + \bY^{(0)})\ .
\end{equation}

Given $\bY$, rather than updating $\bL$, we can instead directly update $\bA$ with any $\calH(\bY)$ proposed in previous SGG works \cite{yang2018graph,yang2021probabilistic,chen2019knowledge}. In this paper, we define the $\calH(\bY)$ as: $[\calH(\bY)]_{ij} \!=\! \bw_a^{T}[\by_i ;  \by_j; \bu_{ij}]$, where $\bw_a \in \mathbb{R}^{3d}$ represents a fusion vector. This enables us to solve the GLD problem, defined in Eq.~\eqref{basic_F}, with a GMP-like procedure as follows:
\begin{equation}\label{mp-l}
\left\{\begin{array}{l}
\tilde\bA^{(k+1)} = {\rm{Normalize}}(\calH(\bY^{(k)})) \\[5pt] 
\bY^{(k+1)} =\frac{1}{3}(\bY^{(k)} + \tilde\bA^{(k+1)} \bY^{(k)} + \bY^{(0)})
\end{array}\right. ,
\end{equation}
where $\tilde\bA=\bD^{-1}\bA $ can be viewed as a row-normalized attention matrix. It is worth noting that nonlinear activation can be incorporated into Eq.~\eqref{mp-l} by solving the revised version of Eq.~\eqref{basic_F} as: $
\calL_{_\text{GLD}}+ \sum\nolimits_{i} \eta\left(\by_{i}\right)
$. Here, $\eta(\by_{i})$ represents an indicator function that assigns infinite penalty to any element of $\by_{i}$ is less than zero. According to the proximal gradient method \cite{li2019lifted}, the proximal descent version of Eq.~\eqref{mp-l} can be written as follows:
\begin{equation}\label{mp-l2}
\left\{\begin{array}{l}
\tilde\bA^{(k+1)} = {\rm{Normalize}}(\calH(\bY^{(k)})) \\[5pt] 
\bY^{(k+1)} ={\rm{ReLU}}(\frac{1}{3}(\bY^{(k)} + \tilde\bA^{(k+1)} \bY^{(k)} + \bY^{(0)}))
\end{array}\right. .
\end{equation}

Regardless of the scalar term (\ie, $\frac{1}{3}$), the only difference, between the GMP layer defined in Eq.~\eqref{mp-ori} and the solver for the GLD problem defined in Eq.~\eqref{mp-l2}, is the skip connection with original node representation $\bY^{(0)}$. Therefore, \textit{existing GMP modules can be utilized as means of solving the GLD problem in SGG}. This conclusion enables us to solve the problem of spurious inter-node correlations in the GLD framework.

\subsubsection{Spuriousness-robust Graph Regularization}
As a quadratic penalty, the Frobenius norm in GLR (Eq.~\eqref{eq:glr}) is known to be sensitive to outliers as errors accumulate quadratically \cite{west1984outlier}. For GMP-based SGG models, this implies that spurious correlations between nodes could dominate the loss, resulting in ambiguous node representations. To address this issue, we propose the following $\ell_p$-based graph regularization to replace GLR in Eq.~\eqref{eq:glr}: 
\begin{equation}\label{pg}
\calG_p (\bY,\bL)=\sum\nolimits_{(i,j) \in \calE}[\bA]_{ij}\kappa_{p}^{\epsilon}\left(\Vert\by_i-\by_j\Vert_2\right). 
\end{equation} 

Accordingly, we can define a general GLD problem as follows:
\begin{equation} \label{GLR_p}
\calL^p_{_\text{GLD}}(\bY,\bL) \triangleq  \left\|\bY - \bX  \right\|_{F}^2 + \calG_p(\bY,\bL).
\end{equation}When $p$ is 2, Eq.~\eqref{GLR_p} is equivalent to Eq.~\eqref{basic_F}, which is the traditional GLD problem. Conventional optimization strategies, \eg, gradient-based or Hessian-based methods, are computationally expensive when optimizing Eq.~\eqref{GLR_p}, especially when $n$ is a large number. Motivated by the majorization-minimization algorithm \cite{sun2016majorization}, we utilize a quadratic upper-bound function to approximate Eq.~\eqref{pg} (Proof is provided in Appendix B). Specifically,
\begin{equation}\label{glrp}
\hat\calG_p(\bY,\bL) ~=~ \sum\nolimits_{(i,j) \in \calE}[\bA]_{ij}[\Omega]_{ij} \Vert\by_i-\by_j\Vert^2_2,
\end{equation}
where
\begin{equation}\label{reweighting1}
[{\Omega}]_{i j} \triangleq\begin{cases}\epsilon^{{p-2}},  &\Vert\by_i-\by_j\Vert_2 \le \epsilon\\ \Vert\by_i-\by_j\Vert^{p-2}_2, & {\rm{otherwise}} \end{cases}.
\end{equation}Here, $[\Omega]_{ij}$ acts as a reweigting factor for $[\bA]_{ij}$. Accordingly, we modify the architecture of U-MP as follows:
\begin{equation}\label{mp-l3}
\left\{\begin{array}{l}
\tilde\bA^{(k+1)} = {\rm{Normalize}}(\Omega^{(k)} \odot \calH(\bY^{(k)})) \\[5pt]
\bY^{(k+1)} ={\rm{ReLU}}(\frac{1}{3}(\bY^{(k)} + \tilde\bA^{(k+1)} \bY^{(k)} + \bY^{(0)}))
\end{array}\right. .
\end{equation}
More details of the U-MP can be found in Appendix C.

Finally, the classification score vector of the $i$-th node can be obtained as follows: $\bt_i={\rm softmax}(\bW_t\hat\by_i)$. Here, $\bW_t \in{\mathbb R}^{O\times d}$ denotes the object classifier, while $\hat\by_i$ is the output node representation obtained by the final U-MP layer.

\subsection{Group Diversity Enhancement}\label{GDE}
Entropy minimization has been widely adopted for optimization in previous SGG models. However, it may also reduce relationship prediction diversity due to issues related to class imbalance and missing annotations; since there are significantly more samples in majority categories, relationship prediction tends to exhibit a bias towards majority categories. In this part, we propose the GDE module to promote relationship prediction diversity. More specifically, the prediction score vector for the relationship between the $i$-th and $j$-th nodes can be expressed as follows:
\begin{equation}\label{rel_f}
\bp_{ij}={\rm{softmax}}(\bW_r(\hat\by_i*\hat\by_j*\bu_{ij})+ {\mbox{\boldmath $f$}}_{ij}),
\end{equation}
where $\bW_r \!\in\! {\mathbb R}^{R\times d}$ denotes the relationship classifier. $*$ denotes a fusion function defined in \cite{tang2019learning}: $\bx * \by = {\mathop{\rm ReLU}\nolimits} \left( {\bW_x\bx + \bW_y\by} \right) - \left( {\bW_x\bx - \bW_y\by} \right) \odot \left( {\bW_x\bx - \bW_y\by} \right)$, where $\bW_x$ and $\bW_y$ project $\bx$, $\by$ to $d$-dimensional space, respectively. ${\mbox{\boldmath $f$}}_{ij}$ indicates the relationship distribution vector between the object categories of the $i$-th and $j$-th nodes in the training set, which functions in the same way as frequency bias and has been widely adopted in existing works \cite{zellers2018neural,tang2019learning,lin2020gps,yang2021probabilistic}. By gathering all prediction score vectors in the same 
image, we obtain a relationship prediction matrix $\bP \in \mathbb{R}^{N\times R}$, which satisfies:
\begin{equation}
\begin{aligned}
&\sum\nolimits_{j=1}^{R} [\bP]_{ij}=1  \ \ \ \\
{\rm s.t.}\ \ [\bP]_{ij} \geq 0, &\ \ \  \forall i \in 1 \ldots N, \ \ \ j \in 1 \ldots R,
\end{aligned}
\end{equation}
where $N$ is the total number of node pairs in the image.

Considering that row-vectors in $\bP$ are linearly independent when predicting different relationship categories, we can utilize the rank of $\bP$ to measure the prediction diversity. However, maximizing the rank of a matrix is known to be an NP-hard problem \cite{sprechmann2015learning}. We propose two strategies to address this issue. 

First, inspired by \cite{liu2018fast,zhang2021fast}, we adopt the $\ell_{2,1}$-norm based regularization to approximate the rank of $\bP$ as follows:
\begin{equation}\label{21}
{\Vert\bP\Vert}_{{{\scriptscriptstyle {2,1}}}}=\sum\nolimits_{j=1}^{R} \sqrt{\sum\nolimits_{i=1}^{N} [\bP]_{ij}^{2}} \ , \end{equation}
which encourages a column-sparse structure for $\bP$, and therefore promotes relationship prediction diversity.

Second, rather than promoting prediction diversity for all node pairs, we find it is more effective to encourage prediction diversity within pairs that share the same object categories. This is mainly because the rank maximization of $\calP$ is hard to optimize when the number of nodes $n$ is large. Accordingly, we divide the node pairs into several groups, each of which contains correlated node pairs. In practice, we find that selecting node pairs of the same object categories for each group is helpful to the optimization of Eq.~\eqref{21}.

Finally, by extending to the whole batch, we can utilize the following loss function to prompt the relationship prediction diversity:
\begin{equation}\label{rp}
{\mathcal{L}}_{e}=\frac{1}{\calM_{B}}{\mathcal{L}}^e_{cls} - \frac{\tau}{B}\sum\nolimits^B_{b=1}\frac{1}{\calN_b}\left\| \bP_b\right\|_{2,1},
\end{equation}
where $\calL^e_{cls}$ denotes the cross-entropy (CE) loss for relationship classification, $\tau$ is a weight, and $B$ represents the number of groups in a mini-batch. Each group contains prediction score vectors for node pairs that share the same object categories. $\bP_b$ denotes the relationship prediction matrix for the $b$-th group, $\calM_{B}$ denotes the number of score vectors in the same batch, while $\calN_b$ represents the number of score vectors in the $b$-th group. 

The critical insight of Eq.~\eqref{rp} is to decrease a certain level of prediction hit rate on majority categories to enhance the prediction hit rate on minority categories. When the prediction diversity increases, one key concern is that some samples belonging to the majority classes may be classified as the minority class. Fortunately, the classification loss on the labeled samples will penalize incorrect predictions caused by encouraging diversity. Consequently, by selecting an appropriate value of $\tau$, the model can generate diverse predictions while ensuring that the vast majority of labeled samples are correctly predicted.

\begin{table*}[t]
\setlength{\tabcolsep}{1.1mm}
\centering
\begin{tabular}{l|l|clclc|c lclc|clclc|l}
\hline
                        &                                       & \multicolumn{5}{c|}{SGDET}                                                                                             & \multicolumn{5}{c|}{SGCLS}                                                                                                         & \multicolumn{5}{c|}{PREDCLS}                                                                                                                      \\
           Backbone                  & Method                               & R@20                   & \multicolumn{1}{c}{} & R@50                   & \multicolumn{1}{c}{} & R@100                  & R@20                   & \multicolumn{1}{c}{}       & R@50                   & \multicolumn{1}{c}{}       & R@100                  & R@20                   & \multicolumn{1}{c}{}       & R@50                   & \multicolumn{1}{c}{}       & R@100    & Mean                                \\ \hline \hline

                          & IMP$^\diamond$  \cite{xu2017scene}                     & 14.6                   &                      & 20.7                   &                      & 24.5                   & 31.7                   &                            & 34.6                   &                            & 35.4                   & 52.7                   &                            & 59.3                   &                            & 61.3    & \multicolumn{1}{c}{39.3}             \\
                          & MOTIFS$^\diamond$  \cite{zellers2018neural}                    & 21.4                   &                      & 27.2                   &                      & 30.3                   & 32.9                   &                            & 35.8                   &                            & 36.5                   & 58.5                   &                            & 65.2                   &                            & 67.1                               &   \multicolumn{1}{c}{43.7}     \\
      & KERN$^\diamond$  \cite{chen2019knowledge}                     & -                      &                      & 27.1                   &                      & 29.8                   & -                      &                            & 36.7                   &                            & 37.4                   & -                      &                            & 65.8                   &                            & 67.6              & \multicolumn{1}{c}{44.1}       \\
 & GPI$^\diamond$  \cite{herzig2018mapping}                     & -                      &                      & -                  &                      & -                   & -                      &                            & 36.5                  &                            & 38.8                  &    -                   &                            & 65.1                 &                           & 66.9           &     \multicolumn{1}{c}{-}       \\

                    & VCTREE$^\diamond$ \cite{tang2019learning}                 & 22.0                   &                      & 27.9                   &                      & 31.3                   & 35.2                   &                            & 38.1                   &                            & 38.8                   & 60.1                   &                            & 66.4                   &                            & 68.1                        &     \multicolumn{1}{c}{45.1}                \\
    \multicolumn{1}{l|}{VGG-16}                          &  GPS-Net$^\diamond$ \cite{lin2020gps} & 22.6 &                      & 28.4 &                      &  31.7 &  36.1 &                            &  39.2 &                            &  40.1 &  60.7 &                            &  66.9 &                            &  68.8 &     \multicolumn{1}{c}{45.9} \\

                                                    &  R-CAGCN$^\diamond$ \cite{yang2021probabilistic} & 22.1 &                      & 28.1 &                      &  31.3 &  35.4 &                            &  38.3 &                            &  39.0&  60.2 &                            &  66.6 &                            &  68.3 &     \multicolumn{1}{c}{45.3}\\
                             &  RelDN$^\ddagger$  \cite{zhang2019graphical} & - &                      &-  &                      &  32.7  &  - &                            &  - &                            &  36.8 &  - &                            &  - &                            &  68.4 &     \multicolumn{1}{c}{-}\\
 
                            &  Seq2Seq-RL$^\ddagger$  \cite{lu2021context} & 22.1 &                      & 30.9 &                      &  34.4  &  34.5 &                            &  38.3 &                            &  39.0 &  60.3 &                            &  66.4 &                            &  68.5 &     \multicolumn{1}{c}{46.3}\\
                            
                                              & RU-Net$^\diamond$                  &\bf 22.9                      & \multicolumn{1}{c}{} &              28.7           & \multicolumn{1}{c}{} &                  32.0      & 37.2                  & \multicolumn{1}{c}{}       & 39.8                   & \multicolumn{1}{c}{}       & 40.9                   & 61.6                  & \multicolumn{1}{c}{}       & 67.8                   & \multicolumn{1}{c}{}       & 69.8                 &     \multicolumn{1}{c}{ 46.6}  \\
                                                    &\bf RU-Net $^\ddagger$                 & 22.6                     & \multicolumn{1}{c}{} &             \bf 31.3           & \multicolumn{1}{c}{} &                 \bf 34.8      &\bf 38.2                  & \multicolumn{1}{c}{}       &\bf 41.2                   & \multicolumn{1}{c}{}       &\bf 42.1                   &\bf 61.9                  & \multicolumn{1}{c}{}       &\bf 68.1                   & \multicolumn{1}{c}{}       &\bf 70.1                 &     \multicolumn{1}{c}{\bf 48.0}  \\\hline

 & VTransE$^*$ \cite{tang2020unbiased}                     & 23.0                   &                      & 29.7                   &                      & 34.3                   & 35.4                   &                            & 38.6                   &                            & 39.4                   & 59.0                   &                            & 65.7                   &                            & 67.6         &  \multicolumn{1}{c}{45.9}            \\

& VCTREE$^*$  \cite{tang2019learning}                         & 24.7                   &                      & 31.5                   &                      & 36.2                   & 37.0                   &                            & 40.5                   &                            & 41.4                   & 59.8                   &                            & 66.2                  &                            & 68.1                                      &   \multicolumn{1}{c}{47.3} \\

 \multicolumn{1}{c|}{RX-101} & MOTIFS$^*$  \cite{zellers2018neural}                      & 25.1                   &                      & 32.1                  &                      & 36.9                   & 35.8                   &                            & 39.1                   &                            & 39.9                   & 59.5                   &                            & 66.0                   &                            & 67.9                              & \multicolumn{1}{c}{47.0}       \\
 
  & SGGNLS$^*$  \cite{zhong2021learning}                         & 24.6                   &                      & 31.8                  &                      & 36.3                   & 36.5                   &                            & 40.0                   &                            & 40.8                   & 58.7                   &                            & 65.6                   &                            & 67.4                              & \multicolumn{1}{c}{47.0}       \\
                          & \bf RU-Net$^*$              &\bf 25.7  &                      & \bf 32.9   &                      &\bf 37.5    & \bf 38.7                   & \multicolumn{1}{c}{}       &\bf 42.4                  & \multicolumn{1}{c}{}       &\bf 43.3                    &\bf 61.2                  & \multicolumn{1}{c}{}       &\bf 67.7                   & \multicolumn{1}{c}{}       &\bf 69.6           & \multicolumn{1}{c}{\bf 48.9}           \\ 
                          \hline
\end{tabular}
\caption{Performance comparisons with state-of-the-art methods on the VG dataset. We compute the mean over all tasks on R@50 and R@100. $^\diamond$, $^\ddagger$, and $^*$ denote using the same Faster-RCNN detector as \cite{zellers2018neural}, \cite{zhang2019graphical}, and \cite{tang2020unbiased}, respectively.}
\label{vg1}
\end{table*}

\subsection{SGG by RU-Net}\label{sgg_i}
During training, the overall loss function $\calL$ for RU-Net can be expressed as follows:
\begin{equation}
\calL=\frac{1}{n_b}\calL^o_{cls}+\calL_e,
\end{equation}
where $n_b$ represents the number of nodes in the batch. $\calL^o_{cls}$ denotes the CE loss for object classification.

During testing, the object category for the $i$-th node is predicted by the following equation:
\begin{equation}
e_i = {\arg{\max} _{o \in {\mathcal O}}}({{\emph {\textbf{\bt}}}_{i}(o)}),
\end{equation}
where $\mathcal O$ represents the set of object categories. The relationship category of the edge between the $i$-th and $j$-th nodes can be obtained as follows:
\begin{equation}
q_{ij}={\arg{\max} _{r \in {\mathcal R}}}({{\emph {\textbf{p}}}_{ij}(r)}),    
\end{equation}
where $\mathcal R$ represents the set of relationship categories.

\begin{table}[t]
\centering
\setlength{\tabcolsep}{1.4mm}
\begin{tabular}{ll|c|c|c}
\hline
& & \multicolumn{1}{c|}{SGDET} & \multicolumn{1}{c|}{SGCLS} & \multicolumn{1}{c}{PREDCLS}  \\
Model & & mR@100 & mR@100 & mR@100 \\ \hline\hline
IMP $^\diamond$\cite{xu2017scene} & & 4.8 & 6.0  & 10.5 \\
FREQ$^\diamond$ \cite{zellers2018neural} && 7.1  & 8.5 & 16.0 \\
MOTIFS $^\diamond$\cite{zellers2018neural} && 6.6  & 8.2 & 15.3 \\
KERN$^\diamond$ \cite{chen2019knowledge}  && 7.3  & 10.0 & 19.2  \\
VCTREE \cite{tang2019learning} && 8.0   & 10.8  & 19.4  \\ 
R-CAGCN$^\diamond$ \cite{yang2021probabilistic} & & 8.8  & 11.1  & 19.9 \\
MOTIFS$^*$ \cite{tang2020unbiased} && 6.8  & 8.5 & 15.8 \\
VCTREE$^*$ \cite{tang2020unbiased} && 6.9   & 7.9  & 16.1  \\ 
Transformer$^*$ \cite{guo2021general} && 8.8   & 10.2  & 17.5  \\ \hline
\bf RU-Net$^\diamond$ & & 10.1  &13.9  &\bf24.7 \\ 
\bf RU-Net$^*$ & &\bf 10.8  &\bf14.6  &24.2 \\ \hline
\end{tabular}
{\caption{Performance comparisons on mean recall $(\%)$ across all 50 relationship categories in the VG dataset.}\label{tbam}}
\end{table}

\begin{figure}[h]
\begin{center}
\includegraphics[scale=0.5]{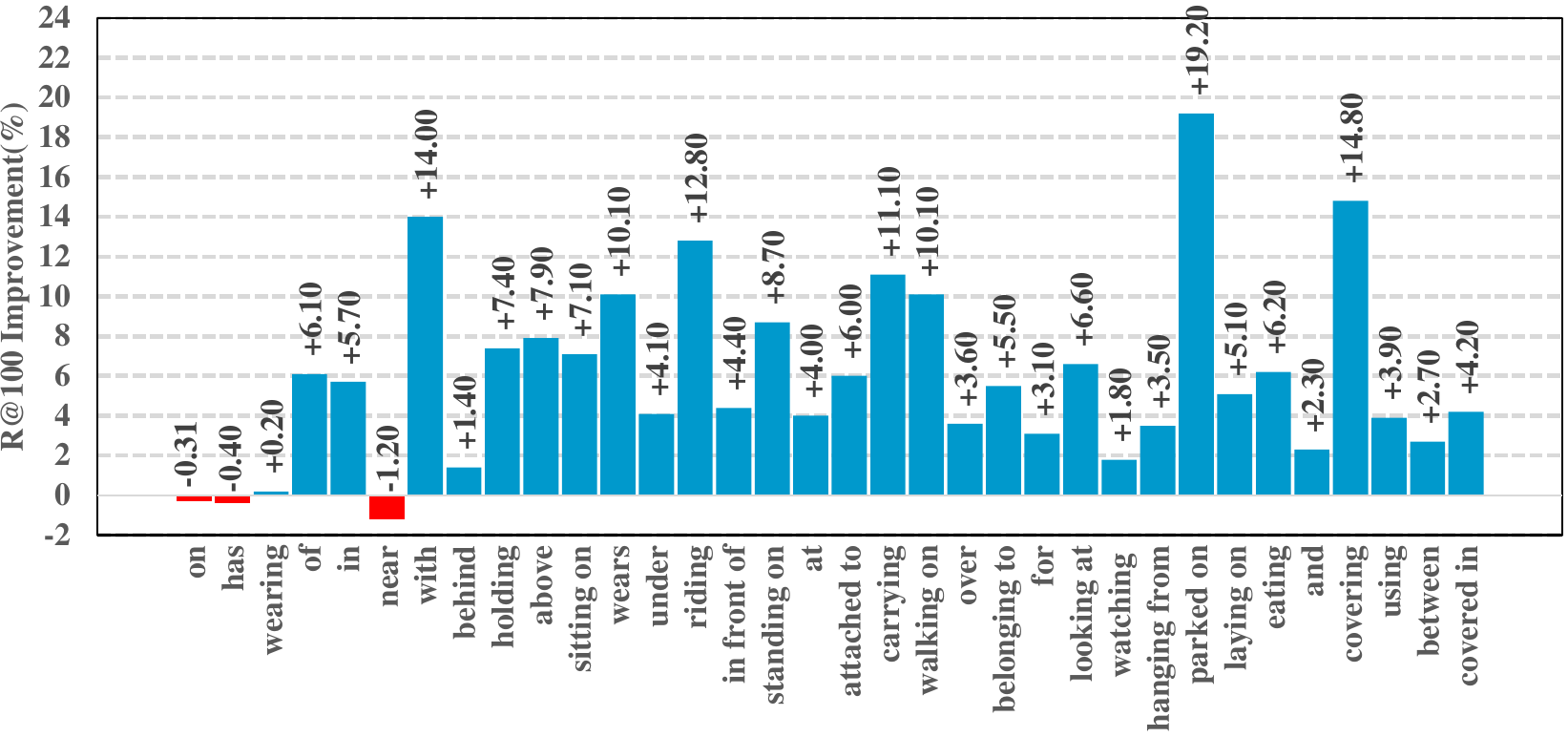}
\caption{Absolute R@100 improvement in PREDCLS by RU-Net compared with R-CAGCN \cite{yang2021probabilistic} on the VG dataset. We use the same backbone and evaluation metric as \cite{tang2019learning}. The Top-35 relationship categories are selected according to their occurrence frequency.}
\label{vg_hist}
\end{center}
\end{figure}

\section{Experiments}
\label{sec:exp}
\subsection{Dataset and Evaluation Settings}
\noindent\textbf{Visual Genome (VG)}: We follow the same data cleaning strategy \cite{xu2017scene} that has been widely used in recent works. The most frequently occurring 150  object categories and 50 relationship categories are utilized for evaluation. We further adopt three conventional protocols for evaluation: (1) Scene Graph Detection (SGDET): Given an image, the model detects objects and predict relationship categories between each pair of objects. (2) Scene Graph Classification (SGCLS): Given the ground-truth location of objects, the model predicts both the object and relationship categories. (3) Predicate Classification (PREDCLS): Given the ground-truth object location and categories, the model predicts only the relationship categories. All algorithms are evaluated using the Recall@$K$ metrics, where $K$=20, 50, and 100, respectively.  Considering that the distribution of relationships in VG is highly imbalanced, we further utilize mean recall@K (mR@$K$) to evaluate the average performance on relationships \cite{chen2019knowledge}.

\noindent\textbf{Open Images (OI)}: We conduct experiments on both Open Images V4 and V6. we follow the same data processing and evaluation protocols utilized in \cite{li2021bipartite,lin2020gps,zhang2019graphical}. The results are evaluated by calculating Recall@50 (R@50), the weighted mean AP of relationships $({\rm {wmAP}}_{rel})$, and the weighted mean AP of phrase (${\rm {wmAP}}_{phr}$). The last metric is given by score$_{wtd}=0.2\times {\rm R}@50+0.4\times {\rm {wmAP}}_{rel}+0.4\times {\rm {wmAP}}_{phr}$. Note that ${\rm {wmAP}}_{rel}$ requires the IoUs between the predicted and ground-truth bounding boxes to be larger than 0.5 for both objects. The ${\rm {wmAP}}_{phr}$ metric is similar but only requires the IoU between the predicted and ground-truth union boxes of the {\textit{subject}} and {\textit{object}} to be over 0.5.

\noindent\textbf{Visual Relationship Detection (VRD)}: We adopt the same dataset split used in \cite{lu2016visual} and the same object detector from \cite{zhang2019graphical}. The evaluation metrics are the same as those in \cite{zhang2019graphical}, which reports R@50 and R@100 for relationship detection and phrase detection, respectively. 

\begin{table}
\setlength{\tabcolsep}{1.3mm}
\centering
\begin{tabular}{ll|c|cc|c}
\hline
\multicolumn{1}{c|}{\multirow{2}{*}{Daraset}} &
  \multicolumn{1}{c|}{\multirow{2}{*}{Model}} &
  \multicolumn{1}{c|}{\multirow{2}{*}{R@50}} &
  \multicolumn{2}{c|}{WmAP} &
  \multicolumn{1}{c}{\multirow{2}{*}{${\rm {score}}_{wtd}$}} \\ \cline{4-5}
\multicolumn{1}{c|}{} &
  \multicolumn{1}{c|}{} &
  \multicolumn{1}{c|}{} &
  rel &
  \multicolumn{1}{l|}{phr} &
  \multicolumn{1}{c}{} \\ \hline \hline
\multicolumn{1}{l|}{}                  & RelDN \cite{zhang2019graphical}      & 74.9                & 35.5                & 38.5                & 44.6                 \\
\multicolumn{1}{c|}{V4}                    & BGNN \cite{li2021bipartite}                                                  & 75.5                & 37.8                & 41.7                 & 46.9                 \\ \cline{2-6} 
\multicolumn{1}{l|}{}                    & \bf{RU-Net}                 & \bf{78.3} &  \bf{38.9} & \bf{42.4} &\bf{48.2}  \\ \hline
\multicolumn{1}{c|}{\multirow{7}{*}{V6}} & RelDN \cite{zhang2019graphical}        & 73.1                & 32.2                & 33.4               & 40.8                 \\
\multicolumn{1}{l|}{}                    & VCTREE \cite{tang2019learning}                                                  & 74.1                & 34.2                & 33.1               & 40.2                 \\
\multicolumn{1}{l|}{}                    & G-RCNN \cite{yang2018graph}                                                  & \multicolumn{1}{c|}{74.5} & \multicolumn{1}{c}{33.2} & \multicolumn{1}{c|}{34.2} & \multicolumn{1}{c}{41.8}  \\
\multicolumn{1}{l|}{}                    & MOTIFS \cite{zellers2018neural}                                                  & \multicolumn{1}{c|}{71.6} & \multicolumn{1}{c}{29.9} & \multicolumn{1}{c|}{31.6} & \multicolumn{1}{c}{38.9}  \\
\multicolumn{1}{l|}{}                    & GPS-Net \cite{lin2020gps}                                                & \multicolumn{1}{c|}{74.8} & \multicolumn{1}{c}{32.9} & \multicolumn{1}{c|}{34.0} & \multicolumn{1}{c}{41.7}  \\
\multicolumn{1}{l|}{}                    & BGNN \cite{li2021bipartite}                                                    & \multicolumn{1}{c|}{75.0} & \multicolumn{1}{c}{33.5} & \multicolumn{1}{c|}{34.2} & \multicolumn{1}{c}{42.1}  \\
\cline{2-6} 
\multicolumn{1}{l|}{}                    & \bf{RU-Net}           & \multicolumn{1}{c|}{\bf 76.9} & \multicolumn{1}{c}{\bf 35.4} & \multicolumn{1}{c|}{\bf 34.9} & \multicolumn{1}{c}{\bf 43.5} \\
\hline
\end{tabular}
\caption{Comparisons with state-of-the-art methods on OI. We adopt the same evaluation metric as in \cite{zhang2019graphical}.}
\label{OI1}
\end{table}

\noindent\textbf{Implementation Details.} 
To facilitate a fair comparison with the majority of existing works, we utilize ResNeXt-101-FPN \cite{lin2017feature, xie2017aggregated} as the backbone for the OI benchmark. We further adopt ResNeXt-101-FPN \cite{lin2017feature, xie2017aggregated} and VGG-16 \cite{simonyan2014very} as the backbones for the VG benchmark. For VRD benchmark, we utilize the VGG-16 \cite{simonyan2014very} as the backbone. During training, we freeze the layers before the ROIAlign layer and optimize the remaining layers in the model using the loss functions described in Section \ref{sgg_i}. We optimize RU-Net via Stochastic Gradient Descent with momentum, using an initial learning rate of 10$^{-3}$ and a batch size of 6. The top-64 object proposals in each image are chosen using per-class non-maximal suppression (NMS) with an IoU of 0.3. Additionally, the sampling ratio between pairs that do not have any relationship (background pairs) and pairs that do have relationships during training is set to 3:1. In all experiments, $\epsilon$ is set to 0.5.

\subsection{Comparisons with State-of-the-art Methods}
\noindent\textbf{Visual Genome:} As Table~\ref{vg1} shows, RU-Net achieves superior performance relative to the current state-of-the-art methods across various metrics. In more detail, RU-Net outperforms the recent GMP-based SGG model, named R-CAGCN \cite{yang2021probabilistic}, by 1.3$\%$ on average at R@50 and R@100 over the three protocols. It also outperforms R-CAGCN \cite{yang2021probabilistic} by 0.7 $\%$, 2.2 $\%$, and 1.5 $\%$ on SGDET, SGCLS, and PREDCLS at Recall@100, respectively. Moreover, RU-Net outperforms VCTREE \cite{tang2019learning} with the same ResNeXt-101-FPN backbone by 1.3$\%$, 1.9$\%$, and 1.5$\%$ on SGCLS, SGDET, and PREDCLS at Recall@100, respectively. Furthermore, to demonstrate RU-Net's robustness to the class imbalance problem on VG, we also compare its performance with state-of-the-art methods using the Mean Recall metric. As shown in Table~\ref{tbam}, RU-Net delivers a notable absolute performance gain, indicating its advantages in handling the class imbalance problem in SGG. To illustrate this advantage more vividly, we present the R@100 improvement of each predicate category compared with R-CAGCN \cite{yang2021probabilistic} under the PREDCLS setting in Figure~\ref{vg_hist}. These improvements are much larger for minority relationship categories. We owe this advantage to the power of the GDE module.

\begin{table}[!t]
\centering
\setlength{\tabcolsep}{1mm}{
\begin{tabular}{lcccccc}
\hline
\multicolumn{1}{l|}{} & \multicolumn{3}{c|}{Relation Detection} & \multicolumn{3}{c}{Phrase Detection} \\
\multicolumn{1}{l|}{Model} & R@50 & \multicolumn{1}{l}{} & \multicolumn{1}{c|}{R@100} & R@50 & \multicolumn{1}{l}{} & R@100 \\ \hline\hline
\multicolumn{1}{l|}{VTransE \cite{zhang2017visual}}& 19.4 &  & \multicolumn{1}{c|}{22.4} & 14.1 &  & 15.2 \\
\multicolumn{1}{l|}{{KL distilation \cite{yu2017visual}}} & 19.2 &  & \multicolumn{1}{c|}{21.3} & 23.1 &  & 24.0 \\
\multicolumn{1}{l|}{Zoom-Net \cite{yin2018zoom}}  & 18.9 &  & \multicolumn{1}{c|}{21.4} & 24.8 &  & 28.1 \\
\multicolumn{1}{l|}{CAI + SCA-M\cite{yin2018zoom}}  & 19.5 &  & \multicolumn{1}{c|}{22.4} & 25.2 &  & 28.9 \\
\multicolumn{1}{l|}{{GPS-Net}\cite{lin2020gps}}  &  21.5 &  &  \multicolumn{1}{c|}{ 24.3} & {28.9} &  & {34.0} \\
\multicolumn{1}{l|}{MF-URLN \cite{zhan2019exploring}}  & 23.9 &  & \multicolumn{1}{c|}{26.8} & 31.5 &  & 36.1 \\
\multicolumn{1}{l|}{RelDN \cite{zhang2019graphical}}  & 25.3 &  & \multicolumn{1}{c|}{28.6} & 31.3 &  & 36.4 \\
\multicolumn{1}{l|}{HetH \cite{wang2020sketching}}  & 22.4 &  & \multicolumn{1}{c|}{24.8} & 30.6 &  & 35.5 \\
\multicolumn{1}{l|}{Seq2Seq-RL \cite{lu2021context}}  & 26.1 &  & \multicolumn{1}{c|}{30.2} & 33.4 &  & 39.1 \\\hline
\multicolumn{1}{l|}{\textbf{RU-Net}}  & \bf 27.4 &  &  \multicolumn{1}{c|}{\bf 31.4} & \textbf{33.8} &  & \textbf{39.5} \\ \hline
\end{tabular}}
\caption{Comparisons with state-of-the-arts on VRD.
}
\label{VRD}
\end{table}

\begin{table}[]
\centering
\setlength{\tabcolsep}{1.55mm}\begin{tabular}{l|cc|cc|cc}
\hline
\multicolumn{1}{l|}{} &\multicolumn{2}{c|}{Module} & \multicolumn{2}{c|}{SGCLS} & \multicolumn{2}{c}{PREDCLS} \\
\multicolumn{1}{c|}{Exp} &\multicolumn{1}{c}{U-MP} & \multicolumn{1}{c|}{GDE} & R@50 &  \multicolumn{1}{c|}{R@100}  & \multicolumn{1}{l}{R@50} &   \multicolumn{1}{l}{R@100} \\ \hline\hline
\multicolumn{1}{c|}{1} &\multicolumn{1}{c}{\xmark}  & \multicolumn{1}{c|}{\xmark} &40.3 & \multicolumn{1}{c|}{41.2} &\multicolumn{1}{c}{{66.0}}  &\multicolumn{1}{c}{{67.8}}\\
\multicolumn{1}{c|}{2} &\multicolumn{1}{c}{\xmark}  & \multicolumn{1}{c|}{\cmark} &40.7 & \multicolumn{1}{c|}{41.6} &\multicolumn{1}{c}{{67.3}}  &\multicolumn{1}{c}{{69.2}}\\
\multicolumn{1}{c|}{3} &\multicolumn{1}{c}{\cmark}  & \multicolumn{1}{c|}{\xmark} &42.2 & \multicolumn{1}{c|}{43.1} &\multicolumn{1}{c}{{66.3}}  &\multicolumn{1}{c}{{68.1}}\\
\hline
\multicolumn{1}{c|}{4} &\multicolumn{1}{c}{\cmark}  & \multicolumn{1}{c|}{\cmark} &\bf42.4 & \multicolumn{1}{c|}{\bf43.3} &\multicolumn{1}{c}{{\bf67.7}}  &\multicolumn{1}{c}{{\bf69.6}}\\
\hline
\end{tabular}
\caption{Ablation studies of the proposed method. We use the same object detection backbone as in \cite{tang2020unbiased}.}
\label{ablation1}
\end{table}

\begin{table*}[t]
\centering
\subfloat[Evaluation on the value of $p$ in Eq.~\eqref{reweighting1}.]{
\tablestyle{3.3 pt}{1.5}\begin{tabular}{ccc|clclclclclcl}
\hline
\multicolumn{3}{c|}{} & \multicolumn{2}{c}{$p$} & \multicolumn{2}{c}{$0$} & \multicolumn{2}{c}{$0.1$} & \multicolumn{2}{c}{$0.3$} & \multicolumn{2}{c}{$1$} & \multicolumn{2}{c}{$2$}\\ \hline\hline
 &  &  & \multicolumn{2}{c}{R@20} & \multicolumn{2}{c}{38.1} & \multicolumn{2}{c}{\bf38.3} & \multicolumn{2}{c}{38.0} & \multicolumn{2}{c}{38.7} & \multicolumn{2}{c}{$37.1$}\\
\multicolumn{3}{c|}{SGCLS} & \multicolumn{2}{c}{R@50} & \multicolumn{2}{c}{41.9} & \multicolumn{2}{c}{\bf42.1} & \multicolumn{2}{c}{41.8} & \multicolumn{2}{c}{41.4}& \multicolumn{2}{c}{$40.8$}\\
\multicolumn{1}{l}{} &  &  & \multicolumn{2}{c}{R@100} & \multicolumn{2}{c}{43.1} & \multicolumn{2}{c}{\bf43.3} & \multicolumn{2}{c}{43.0} & \multicolumn{2}{c}{42.6} & \multicolumn{2}{c}{$41.9$}\\ \hline
\end{tabular}}\hspace{0.3mm}
\subfloat[Evaluation on the number of U-MP layers $K$.]{
\tablestyle{4.1 pt}{1.5}\begin{tabular}{ccc|clclclclcl}
\hline
\multicolumn{3}{c|}{} & \multicolumn{2}{c}{$K$} & \multicolumn{2}{c}{2} & \multicolumn{2}{c}{3} & \multicolumn{2}{c}{4} & \multicolumn{2}{c}{5} \\ \hline\hline
 &  &  & \multicolumn{2}{c}{R@20} & \multicolumn{2}{c}{38.1} & \multicolumn{2}{c}{38.4} & \multicolumn{2}{c}{38.5} & \multicolumn{2}{c}{\bf 38.7} \\
\multicolumn{3}{c|}{SGCLS} & \multicolumn{2}{c}{R@50} & \multicolumn{2}{c}{41.8} & \multicolumn{2}{c}{42.0} & \multicolumn{2}{c}{42.2} & \multicolumn{2}{c}{\bf 42.4} \\
\multicolumn{1}{l}{} &  &  & \multicolumn{2}{c}{R@100} & \multicolumn{2}{c}{42.7} & \multicolumn{2}{c}{43.0} & \multicolumn{2}{c}{43.1} & \multicolumn{2}{c}{\bf 43.3} \\ \hline
\end{tabular}}\hspace{0.3mm}
\subfloat[Evaluation on the value of $\tau$ in Eq.~\eqref{rp}.]{
\tablestyle{3.3 pt}{1.5}\begin{tabular}{ccc|clclclclcl}
\hline
\multicolumn{3}{c|}{} & \multicolumn{2}{c}{$\tau$} & \multicolumn{2}{c}{$0.05$} & \multicolumn{2}{c}{$0.1$} & \multicolumn{2}{c}{$0.15$} & \multicolumn{2}{c}{$0.2$}\\ \hline\hline
 &  &  & \multicolumn{2}{c}{R@20} & \multicolumn{2}{c}{60.8} & \multicolumn{2}{c}{\bf61.2} & \multicolumn{2}{c}{60.5} & \multicolumn{2}{c}{$59.9$}\\
\multicolumn{3}{c|}{PREDCLS} & \multicolumn{2}{c}{R@50} & \multicolumn{2}{c}{67.3} & \multicolumn{2}{c}{\bf67.7} & \multicolumn{2}{c}{67.0} & \multicolumn{2}{c}{$66.4$}\\
\multicolumn{1}{l}{} &  &  & \multicolumn{2}{c}{R@100} & \multicolumn{2}{c}{69.2} & \multicolumn{2}{c}{\bf69.6} & \multicolumn{2}{c}{68.9}& \multicolumn{2}{c}{$68.3$} \\ \hline
\end{tabular}}\hspace{1mm}
\vspace{-2mm}
{\caption{The impact of hyperparameters on the U-MP and GDE modules, respectively.}\label{ablation_whole}}
\vspace{-2mm}
\label{tab:2}
\end{table*}

\begin{table}[t]
\centering
\tablestyle{4 pt}{1.5}\begin{tabular}{ccc|clclclcl}
\hline
\multicolumn{3}{c|}{} & \multicolumn{2}{c}{Group} & \multicolumn{2}{c}{$\mathcal{I}$} & \multicolumn{2}{c}{$\mathcal{B}$+$\mathbb{G}$} & \multicolumn{2}{c}{$\mathcal{B}$+$\mathbb{G}^{*}$} \\ \hline\hline
\multicolumn{1}{l}{} &  &  & \multicolumn{2}{c}{R@50} & \multicolumn{2}{c}{66.7} & \multicolumn{2}{c}{67.2} & \multicolumn{2}{c}{\bf67.7} \\
\multicolumn{3}{c|}{PREDCLS}&\multicolumn{2}{c}{R@100} & \multicolumn{2}{c}{68.5} & \multicolumn{2}{c}{69.1} & \multicolumn{2}{c}{\bf69.6} \\ \cline{4-11}
\multicolumn{1}{l}{} &  &  & \multicolumn{2}{c}{mR@100} & \multicolumn{2}{c}{22.5} & \multicolumn{2}{c}{23.8} & \multicolumn{2}{c}{\bf24.2} \\ \hline
\end{tabular}
\vspace{-2mm}
{\caption{The Design Choices for the GDE modules.}\label{GDE_a}}
\vspace{-4mm}
\end{table}

\noindent\textbf{Open Images:} 
We compare the performance of RU-Net with state-of-the-art methods in Table~\ref{OI1}. Using the same object detector, RU-Net outperforms RelDN \cite{zhang2019graphical} by 3.6$\%$ and 2.7$\%$ in terms of overall metric score$_{wtd}$ for OI V4 and V6, respectively. More specifically, in OI V4, RU-Net outperforms RelDN by 3.4$\%$, 3.4$\%$, and 3.9$\%$ on R@50, wmAP$_{rel}$, and wmAP$_{phr}$, respectively. Furthermore, when compared with other approaches for OI V6, RU-Net consistently achieves the best performance.

\noindent\textbf{Visual Relationship Detection:} 
In Table~\ref{VRD}, we compare the performance of RU-Net with state-of-the-art methods on the VRD dataset. It can be seen that RU-Net consistently achieves superior performance under both relation detection and phrase detection metrics.

\subsection{Ablation Studies}\label{ablation_all}
\noindent\textbf{Effectiveness of the Proposed Modules.} We first perform an ablation study to justify the effectiveness of U-MP and GDE. The results are summarized in Table \ref{ablation1}. Exp 1 in Table~\ref{ablation1} shows the performance of the baseline, which adopt neither U-MP or GDE modules. It employs the GMP module defined in Eq.~\eqref{mp-ori} for message passing. To facilitate fair comparison, all the other settings remain the same as RU-Net. Exps 2-4 show that each module helps to promote the performance of SGG. The best performance is achieved when both modules are involved. Note that U-MP and GDE are designed to refine object and relationship representations, respectively. Therefore, U-MP helps the model achieve outstanding SGCLS performance, which heavily depends on the object classification ability. Meanwhile, GDE enables the model to achieve a significant performance gain on the PREDCLS task, mainly relying on relationship prediction power.

\noindent\textbf{Evaluation on hyperparameters for U-MP and GDE.}
We go on to verify the impact of the hyperparameters of the U-MP and GDE modules. As shown in Table~\ref{ablation_whole}(a), RU-Net achieves the best performance when $p$ is set to 0.1 in the $\ell_p$-based graph regularization. In Table~\ref{ablation_whole}(b), we show the performance of RU-Net with different numbers of U-MP layers, ranging from two to five. The model performance improves consistently as the number of U-MP layers increases. However, due to limitations on GPU memory size, we only conduct experiments up to five U-MP layers. Finally, the value of the weight $\tau$ determines the impact of the $\ell_{2,1}$-based regularization on relationship prediction. As shown in Table~\ref{ablation_whole}(c), the model achieves the best performance when $\tau$ equals 0.1.

\noindent\textbf{Design Choices for the GDE module.} In Table~\ref{GDE_a}, we compare the performance of GDE with and without the grouping strategy described in Eq.~\eqref{rp}. ``$\mathcal{I}$" denotes that we impose the diversity regularization on relationship predictions for each training image. ``$\calB+\mathbb{G}$" represents that we divide all node pairs in batch into several groups according to the object categories of the nodes. For each group, we impose an $\ell_{2,1}$-based regularization term. Besides, ``$\calB+\mathbb{G}^*$" means we remove small groups that contain less than three elements.  Experimental results in Table~\ref{GDE_a} show that the grouping strategy consistently achieves better performance.

\subsection{Conclusion and Limitations}
In this paper, we propose the RU-Net model, which adopts scene graph-based regularizations to handle two critical issues in SGG: ambiguous node representations and low relationship prediction diversity. From the perspective of the unrolling technique, we first prove that GMP can be interpreted as a solver for GLD. We then address the ambiguous node representation problem with the U-MP module, which utilizes an $\ell_p$-based graph regularization to suppress spurious correlations between nodes. We further enhance the diversity in the relationship prediction through a group-wise $\ell_{2,1}$-based regularization term. Extensive experimental results justify the effectiveness of RU-Net on three popular SGG datasets. Like most SGG models, one limitation of our method is its dependency on pre-trained object detectors~\cite{ren2015faster,wang2021fp}. In the future, we will apply the proposed techniques to end-to-end SGG models. We hope this study will provide valuable insights for future research to design interpretable and robust SGG models.

\noindent\textbf{Broader Impacts.}
SGG is able to simultaneously provide object and relationship predictions. This merit enables more in-depth scene understanding and can potentially benefit many real-world applications, like intelligent service robot and autonomous driving. We do not foresee any negative societal consequences arising specifically from our contributions in this paper.

\noindent{\textbf {Acknowledgment}}. This work is supported by the NSF of China (Nos: 62076101, 62002090), the Program for Guangdong Introducing Innovative and Entrepreneurial Teams (No: 2017ZT07X183), and the ARC FL-170100117.

{\small
\bibliographystyle{ieee_fullname}
\bibliography{egbib}
}

\end{document}